\documentclass[3p,times,procedia]{elsarticle}
\usepackage{ecrc}
\usepackage[bookmarks=false]{hyperref}
    \hypersetup{colorlinks,
      linkcolor=blue,
      citecolor=blue,
      urlcolor=blue}
\volume{00}

\firstpage{1}

\journalname{Procedia Computer Science}

\runauth{Jiaqiang Zhang et al.}

\jid{procs}

\usepackage{booktabs}
\usepackage{graphicx}%
\usepackage{multirow}%
\usepackage{amsmath,amssymb,amsfonts}%
\usepackage{booktabs}%
\usepackage{algorithm}%
\usepackage{algorithmicx}%
\usepackage{algpseudocode}%
\usepackage{listings}%
\usepackage{subcaption}
\usepackage{standalone}
\usepackage{tikz}
\usepackage{pgfplots}
\pgfplotsset{compat=1.7}
\usepgfplotslibrary{groupplots}

\newlength\figureheight
\newlength\figurewidth
\usepackage[figuresright]{rotating}
\begin{document}
\begin{frontmatter}



\dochead{The 17th International Conference on Ambient Systems, Networks and Technologies (ANT) \\ April 14-16, 2026, Istanbul, Türkiye}%
%% e.g. \dochead{17th International Conference on Dynamical Processes in Excited States of Solids}

\title{Seamless Outdoor–Indoor Pedestrian Positioning System with GNSS/UWB/IMU Fusion: A Comparison of EKF, FGO, and PF
% Seamless Outdoor–Indoor Pedestrian Positioning System with GNSS/UWB/IMU Fusion: A Comparative Analysis of FGO, EKF, and PF
}


\author[a]{Jiaqiang Zhang\corref{cor1}} 
\author[a]{Xianjia Yu}
\author[a]{Sier Ha}
\author[a]{Paola Torrico Moron}
\author[a]{Sahar Salimpour}
\author[a]{Farhad Kerama}
\author[a]{Haizhou Zhang}
\author[a]{Tomi Westerlund}

\address[a]{\href{https://tiers.utu.fi}{Turku Intelligent Embedded and Robotic Systems (TIERS) Lab, University of Turku, Finland}}

\begin{abstract}
% Accurate pedestrian positioning across outdoor-indoor environments remains challenging because available sensors are complementary yet individually limited: GNSS is only reliable in open outdoor environments but degrades drastically in urban canyons, UWB offers precise local indoor coverage but limited range, and inertial PDR provides continuous motion estimates that drift over time. This paper presents a probabilistic fusion framework that integrates GNSS, UWB, and chest-mounted IMU-PDR within an error-state Extended Kalman Filter, augmented by map-based environmental constraints derived from OpenStreetMap, with GUI visualization provided. Building footprints are used to enforce physical feasibility by rejecting or projecting implausible GNSS fixes and constraining the fused trajectory during outdoor–indoor transitions. The system is implemented on a wearable Raspberry Pi 5 platform equipped with GNSS, IMU, and UWB sensors, running in real time under ROS 2 with visualization in Foxglove. Quantitative evaluation in an urban campus and factory-like indoor environment—using RTK and optical motion capture as references—demonstrates that the proposed approach achieves seamless, drift-bounded pedestrian tracking and smooth handover between GNSS- and UWB-dominant regimes. The results confirm that lightweight map constraints can effectively regularize multi-sensor fusion for robust pedestrian navigation across complex mixed environments.

Accurate and continuous pedestrian positioning across outdoor–indoor environments remains challenging because GNSS, UWB, and inertial PDR are complementary yet individually fragile under signal blockage, multipath, and drift. This paper presents a unified GNSS/UWB/IMU fusion framework for seamless pedestrian localization and provides a controlled comparison of three probabilistic back-ends: an error-state extended Kalman filter, sliding-window factor graph optimization, and a particle filter. The system uses chest-mounted IMU-based PDR as the motion backbone and integrates absolute updates from GNSS outdoors and UWB indoors. To enhance transition robustness and mitigate urban GNSS degradation, we introduce a lightweight map-based feasibility constraint derived from OpenStreetMap building footprints, treating most building interiors as non-navigable while allowing motion inside a designated UWB-instrumented building. The framework is implemented in ROS 2 and runs in real time on a wearable platform, with visualization in Foxglove. We evaluate three scenarios: indoor (UWB+PDR), outdoor (GNSS+PDR), and seamless outdoor–indoor (GNSS+UWB+PDR). Results show that the ESKF provides the most consistent overall performance in our implementation.
\end{abstract}

\cortext[cor1]{Corresponding author.}
\end{frontmatter}

%\correspondingauthor[*]{Corresponding author. Tel.: +0-000-000-0000 ; fax: +0-000-000-0000.}
\email{jiaqiang.zhang@utu.fi}

% \linenumbers


%\enlargethispage{-7mm}
%%%%%%%%%%%%%%%%%%%%%%%%%%%%%%%%%%%%%%%%%%%%%%
%%                                          %%
%%              INTRODUCTION                %%
%%                                          %%
%%%%%%%%%%%%%%%%%%%%%%%%%%%%%%%%%%%%%%%%%%%%%%

% \textbf{NOTE:} use \red{red for TODOs / work in progress} and use \blue{blue for done / almost done paragraphs}
% use \hl{} for hightlight
% textbf{This should be bold!}
\section{Introduction}
\label{sec:introduction}
Accurate and continuous pedestrian positioning across outdoor and indoor environments remains challenging due to the complementary but individually limited nature of available sensors. Outdoors, Global Navigation Satellite System (GNSS) provides globally referenced positions but suffers in urban areas from signal blockage, multipath, and dilution of precision near buildings. Indoors, GNSS is largely unavailable; Ultra-Wideband (UWB) can deliver decimeter-level accuracy where anchors are deployed, but offers only local coverage and is sensitive to non-line-of-sight (NLOS) bias. Inertial sensing enables self-contained motion tracking at a high rate, yet pure integration drifts over time. In practice, maintaining seamless accuracy as a user traverses doorways, corridors, and urban canyons requires principled fusion of these modalities together with environmental knowledge that constrains physically plausible motion.

A broad spectrum of probabilistic estimators has been applied to multi-sensor pedestrian positioning. Extended Kalman filtering (EKF), especially in its error-state form, is attractive for real-time embedded systems due to its computational efficiency and structured treatment of high-rate inertial updates. Factor graph optimization (FGO) offers a smoothing perspective that can incorporate heterogeneous constraints over a window, often improving global consistency and robustness to intermittent measurement availability. Particle filtering (PF) provides a flexible nonparametric alternative capable of representing multimodal posteriors and handling strong nonlinearities or non-Gaussian errors that frequently arise in GNSS multipath and UWB NLOS conditions. Yet, for seamless outdoor--indoor pedestrian positioning using the same GNSS/UWB/IMU stack, a carefully controlled comparison of these three estimator families under consistent sensing, map constraints, and evaluation protocols remains limited.

\begin{figure}[t]
    \centering
    \includegraphics[width=0.8\textwidth, trim=20 40 20 30, clip]{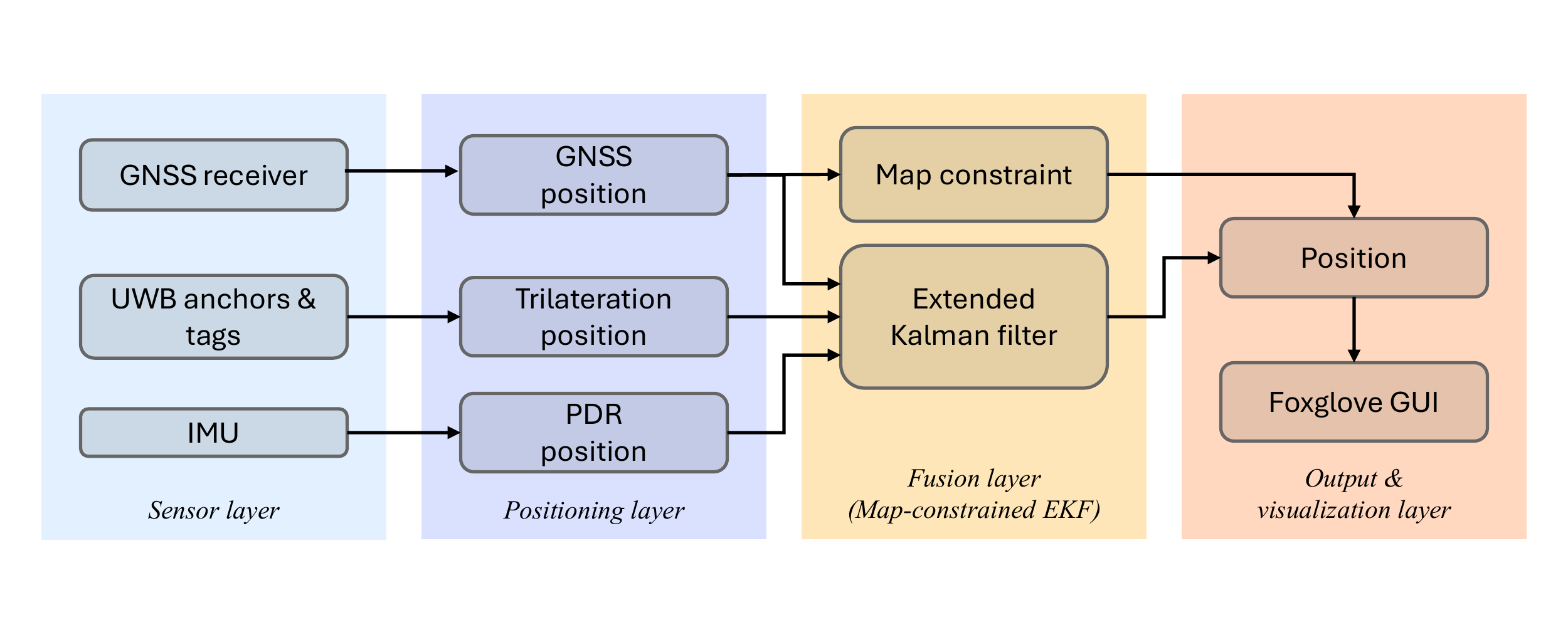}
    \caption{Proposed evaluation scheme in this study}
    \label{fig:system_arch}
    \vspace{-0.8cm}
\end{figure}

This work addresses seamless outdoor--indoor pedestrian positioning by combining GNSS, UWB, and Pedestrian Dead Reckoning (PDR) from a chest-mounted IMU within three probabilistic fusion frameworks augmented by map constraints: an error-state EKF (ESKF), an FGO-based sliding-window smoother, and a PF-based estimator. We target scenarios such as industrial sites or sensitive buildings in urban settings, where users move intermittently between open-sky areas and factory-like interiors. Our key observation is that coarse environmental structure---specifically, building footprints---can be exploited online to (i) reject or down-weight physically impossible GNSS fixes (e.g., spuriously placing a user inside a building without UWB coverage) and (ii) regularize the fused trajectory near transitions, reducing discontinuities and large outliers. By integrating the same map knowledge into all three estimators, we aim to isolate algorithmic differences from environmental priors.

The proposed system is realized on a wearable platform comprising a Raspberry Pi,5, a GNSS module, an Xsens MTi-300 IMU, and a Decawave DWM1001 UWB tag mounted on a chest strap. PDR is derived from the IMU by detecting steps and estimating per-step displacements and heading increments; these kinematic increments provide drift-resistant relative motion over walking horizons and supply high-rate constraints between absolute updates. UWB ranges to surveyed anchors inside a designated building yield accurate indoor updates, while GNSS provides absolute position outdoors. In the ESKF, we (i) treat PDR increments as control-like motion updates, (ii) loosely couple GNSS position measurements, and (iii) tightly incorporate UWB range observations. In the FGO formulation, PDR increments, GNSS positions, and UWB ranges are expressed as factors over a sliding window, enabling joint refinement of recent states and improved consistency during brief GNSS outages or UWB geometry changes. In the PF, PDR increments define the proposal dynamics, while GNSS and UWB likelihoods re-weight particles; this framework allows the posterior to remain multi-hypothesis in ambiguous transition zones.

To inject environmental knowledge, we query OpenStreetMap via Overpass Turbo to obtain building polygons (Fig.~\ref{fig:building_polygons}). At runtime, all building interiors are treated as forbidden regions for the fused solution except the single building equipped with UWB anchors, which is designated as free space. The map constraint is implemented consistently across estimators: in the ESKF by rejecting or projecting implausible GNSS updates to the nearest admissible boundary (Fig.~\ref{fig:map_constraint_demo}); in the FGO by adding feasibility-aware penalties or gating factors that discourage solutions inside forbidden regions; and in the PF by assigning near-zero likelihood to particles that violate the occupancy rule. This unified treatment prevents physically impossible corrections and reduces multipath-induced excursions along facades, while still permitting motion within the UWB-covered interior.

All data are acquired and synchronized using ROS~2, and we provide real-time GUI visualization in Foxglove to display trajectories and map overlays. The system is quantitatively evaluated in an urban campus with a factory-like indoor lab: outdoors, ground truth is provided by an Xsens MTi-680G with Real Time Kinematic (RTK); indoors, by an optical motion-capture system aligned to the global frame. These complementary references allow continuous accuracy assessment across environment transitions and support a fair comparison of ESKF, FGO, and PF under outdoor, indoor, and outdoor--indoor sequences.

This paper aims to: (i) design a map-constrained fusion pipeline that integrates GNSS, UWB, and IMU-PDR to yield robust, drift-bounded pedestrian trajectories across outdoor--indoor transitions; (ii) implement three representative estimators---ESKF, FGO, and PF---under consistent sensing and constraint assumptions; and (iii) provide an empirical comparison that clarifies when the efficiency of filtering, the global consistency of smoothing, or the flexibility of nonparametric inference offers the greatest benefit for seamless pedestrian positioning.

We make four contributions:
(1) a unified GNSS/UWB/IMU-PDR fusion architecture with a consistent building-footprint constraint applied across ESKF, FGO, and PF;
(2) a lightweight, real-time use of open-source map polygons to enforce feasibility while permitting motion within a designated UWB-covered building;
(3) a reproducible ROS,2/Foxglove implementation on an embedded Raspberry Pi,5 with commodity sensors and three estimator back-ends; and
(4) a dual-ground-truth evaluation protocol enabling fine-grained assessment of accuracy and continuity during doorway crossings and urban canyon segments.
We assume a single permitted building equipped with surveyed UWB anchors, treat all other building interiors as non-navigable, and focus on horizontal positioning for a walking pedestrian; vertical motion and multi-floor transitions are left to future work.

The rest of the paper is organized as follows.
Section~\ref{sec:related_work} provides an overview of related work on GNSS/UWB/IMU fusion and outdoor--indoor transition handling.
Section~\ref{sec:methodology} details the system architecture, PDR modeling, measurement models, map-constraint mechanisms, and the three estimator formulations.
Section~\ref{sec:experiment} presents results for outdoor, indoor, and outdoor--indoor scenarios and compares the ESKF, FGO, and PF performance.
% Section~\ref{sec:discussion} discusses limitations and practical implications.
Finally, Section~\ref{sec:conclusion} concludes the paper.

%%%%%%%%%%%%%%%%%%%%%%%%%%%%%%%%%%%%%%%%%%%%%%
%%                                          %%
%%              RELATED WORKS               %%
%%                                          %%
%%%%%%%%%%%%%%%%%%%%%%%%%%%%%%%%%%%%%%%%%%%%%%

% \newpage
\section{Related Work}
\label{sec:related_work}

Seamless pedestrian localization increasingly fuses absolute outdoor positioning with infrastructure-aided and inertial methods indoors, often augmented by environment awareness or explicit constraints. Multi-sensor fusion is a representative direction: Zhang et al.~\cite{zhang2025seamless} integrate GNSS/IMU/UWB with map information to mitigate multipath and stabilize trajectories across indoor–outdoor transitions, closely aligned with our use of building polygons as feasibility constraints. A low-cost wearable platform that loosely couples GNSS and UWB with an Inertial Navigation System (INS) backbone demonstrated continuity at indoor–outdoor boundaries in~\cite{di2020loosely}. Robust estimator design for UWB/PDR integration has also been explored: an adaptive EKF with an additional heading constraint improves resilience to UWB non-line-of-sight and reduces PDR drift in~\cite{yuan2021robustly}. On commodity devices, environment-perception-based switching orchestrates GNSS/INS/BLE for smooth handovers near doorways~\cite{liu2022environment}, and BLE-aided particle-filter fusion yields real-time indoor positioning when combined with PDR~\cite{jin2023real}. A dynamic feasible-region IMU/UWB fusion method restricts the solution space to improve robustness in cluttered interiors~\cite{liu2024dynamic}.

Within inertial-centric pedestrian dead reckoning, recent work prioritizes heading stability and drift mitigation for general-purpose wearables and unconstrained device poses. A multi-source PDR approach fusing accelerometer, gyroscope, geomagnetic, and ancillary cues improves gait detection, step length, and heading estimation indoors~\cite{wu2024indoor}; complementary yaw-correction with two low-cost MIMUs further suppresses heading drift and improves downstream position estimates~\cite{wang2023yaw}. For smartphone deployments, Jiang et al. implement PDR/GNSS integration using both a Kalman filter and factor-graph optimization—modeling PDR as step-wise factors and GNSS as absolute position factors—and show that factor-graph fusion outperforms KF-based fusion in field trials~\cite{jiang2022implementation}. A recent overview synthesizes model-based and data-driven pedestrian inertial navigation, highlighting trends in learned PDR, hybrid fusion, and the role of priors and constraints for wearables~\cite{klein2025pedestrian}.

Body-mounted wearables that centralize sensors on the torso or head illustrate how PDR priors, visual/radio aiding, and learned models can be combined for robust tracking. A chest-mounted localization vest fuses advanced adaptive PDR (dynamic step segmentation and adaptive step length) with binocular vision in an EKF, yielding substantial accuracy gains over vision-only baselines across diverse indoor motions~\cite{tian2025autonomous}. In parallel, ROCIP employs a learned probabilistic motion model with a Rao–Blackwellized particle filter to maintain bounded inertial drift during complex human–environment interactions~\cite{hou2025rocip}. A recent fusion system integrating UWB, IMU, and vision further underscores the benefits of multi-sensor opportunism for indoor robustness~\cite{deng2025fusion}. Taken together with map-aware GNSS/IMU/UWB fusion~\cite{zhang2025seamless} and feasible-region IMU/UWB constraints~\cite{liu2024dynamic}, these threads suggest a design space in which wearable PDR provides high-rate kinematic increments, absolute anchors (GNSS/UWB) provide global consistency, and environmental constraints regularize the estimator near building boundaries—an approach we adopt within an ESKF.

\section{Methodology}
\label{sec:methodology}

\subsection{Proposed Pedestrian Positioning System}
\label{subsec:proposed}
The proposed framework for this study can be found in Fig.~\ref{fig:system_arch}. 
A chest-mounted platform provides step-wise PDR from an Xsens MTi-300 IMU, while absolute position updates are obtained from GNSS outdoors and UWB-based trilateration indoors.
All measurements and estimates are expressed in a local ENU frame anchored at the test site.

To compare filtering and smoothing paradigms under the same sensing conditions, we implement three back-ends: a planar error-state EKF, a sliding-window factor graph optimizer, and a particle filter. The three methods share the same PDR front-end and consume the same absolute position topics (GNSS fix and UWB trilateration fix).

% The estimator is a planar error-state Extended Kalman Filter (ES-EKF). The PDR front-end provides high-rate kinematic prediction updates, while GNSS positions or UWB-based trilateration supply absolute position corrections whenever available. To inject environmental knowledge, we derive a feasible region from building polygons obtained via Overpass Turbo (OpenStreetMap). All building interiors are treated as forbidden, except a single allowed building that is instrumented with UWB anchors for indoor experiments.

% The estimator is prevented from entering forbidden areas using two complementary mechanisms: (i) GNSS measurement pre-filtering, which rejects or projects GNSS fixes that would place the pedestrian inside disallowed buildings; and (ii) post-update projection of the EKF state to the nearest feasible point whenever the updated position leaves the feasible region. This combination enforces physically plausible trajectories across outdoor segments, indoor segments, and indoor–outdoor transitions, while preserving the probabilistic nature of the filter.

% \subsubsection{Chest-Mounted Pedestrian Dead Reckoning}
% \label{subsubsec:pdr}
The Chest-Mounted PDR front-end runs at an effective rate of 50~Hz and outputs step-triggered horizontal
displacement increments in ENU.
Steps are detected from the band-pass filtered dynamic acceleration magnitude within
the typical walking frequency range.
Step length is estimated using a Weinberg-style relation driven by the peak--trough amplitude, where the peak--valley difference is defined as $\Delta A_k = A^{\text{pk}}_k - A^{\text{val}}_k$, and the resulting step length is computed as $\Delta s_k = k\,(\Delta A_k)^{1/4}$, with $k$ calibrated from a short walk.
The step displacement used by all fusion back-ends is
% \vspace{-20pt}
\begin{equation}
\Delta \mathbf{p}_k =
\begin{bmatrix}
\Delta s_k \sin(\psi_k) \\
\Delta s_k \cos(\psi_k)
\end{bmatrix},
\end{equation}
% \vspace{-20pt}

with the heading $\psi_k$ obtained from the IMU/magnetometer-based orientation estimate.
This shared PDR design ensures that performance differences mainly originate from the
probabilistic back-ends rather than the motion front-end.

GNSS provides low-rate global fixes outdoors, while indoor absolute positions are produced by
a UWB trilateration node and delivered to the fusion back-ends as \texttt{NavSatFix}.
Each absolute measurement is transformed into ENU for fusion. 
In the subsequent evaluation, GNSS and UWB updates are selectively used according to their
availability per scenario. 

To provide a fair and compact comparison under identical sensing conditions, we implement three probabilistic back-ends that share the same chest-mounted PDR front-end and consume the same absolute position fixes from GNSS (outdoor) and UWB trilateration (indoor). The ESKF operates as a lightweight recursive filter: instead of integrating raw inertial acceleration, it treats the PDR displacement as the primary motion input, and uses sporadic GNSS/UWB fixes as position corrections to bound drift. This design yields high computational efficiency and stable real-time behavior when the error distribution is approximately unimodal. In contrast, the sliding-window FGO formulates the problem as a local MAP smoothing task over a compact pose chain. Consecutive nodes are connected by PDR between-factors, while GNSS/UWB updates are injected as absolute position factors; a weak anchor and window pruning maintain numerical stability and limit computational growth. This allows later absolute information to retroactively refine recent steps, which is beneficial during intermittent updates or transitions. Finally, the PF represents the posterior by a set of weighted particles and propagates them only upon detected steps using the same PDR increments with stochastic perturbations. Particle weights are updated by the GNSS/UWB likelihood, and resampling is triggered when degeneracy is detected via the effective sample size. This sampling-based formulation is more tolerant to non-Gaussian noise and occasional outliers, at the cost of higher computational demand. Together, these three back-ends cover filtering, smoothing, and sampling perspectives while keeping the motion and measurement interfaces consistent for systematic evaluation across outdoor, indoor, and transition scenarios.

\begin{figure*}[t]
\centering
% \scalebox{0.9}{%
% ===================== TOP ROW (2 blocks) =====================
\begin{minipage}[t]{0.47\textwidth}
\begin{algorithm}[H]
% \captionsetup{type=algorithm}
\caption{PDR-driven ESKF}
\label{alg:esekf_invpin}
\begin{algorithmic}[1]
\State \textbf{State:} $(\mathbf{p}_{\text{est}},\mathbf{v}_{\text{est}},q_{\text{est}},\mathbf{P})$
\State \textbf{Error:} $[\delta\mathbf{p},\delta\mathbf{v},\delta\boldsymbol{\theta}]$, $\mathbf{P}\in\mathbb{R}^{9\times 9}$
\State Init from first accepted absolute fix
\For {each IMU/PDR sample}
  \State $\Delta\mathbf{p}\!\gets\! \mathbf{p}^{\text{PDR}}-\mathbf{p}^{\text{PDR}}_{\text{prev}}$
  \State $\mathbf{p}_{\text{est}}\!\gets\!\mathbf{p}_{\text{est}}+\Delta\mathbf{p}$;\; $q_{\text{est}}\!\gets\! q^{\text{PDR}}$
  \State $\mathbf{P}\!\gets\!\mathbf{F}\mathbf{P}\mathbf{F}^\top+\mathbf{L}\mathbf{Q}\mathbf{L}^\top$
\EndFor
\For {each absolute fix (GNSS/UWB)}
  \State $\mathbf{z}\!\gets\!\mathrm{geodetic2enu}(\mathbf{z}_{\text{LLA}})$
  \State $\mathbf{K}\!\gets\!\mathbf{P}\mathbf{H}^\top(\mathbf{H}\mathbf{P}\mathbf{H}^\top+\sigma_{\text{sensor}}^2\mathbf{I})^{-1}$
  \State $\delta\mathbf{x}\!\gets\!\mathbf{K}(\mathbf{z}-\mathbf{p}_{\text{est}})$
  \State $\mathbf{p}_{\text{est}}\!\leftarrow\!\mathbf{p}_{\text{est}}+\delta\mathbf{p}$;\; $\mathbf{v}_{\text{est}}\!\leftarrow\!\mathbf{v}_{\text{est}}+\delta\mathbf{v}$
  \State $q_{\text{est}}\!\leftarrow\!\mathrm{Exp}(\delta\boldsymbol{\theta})\!\otimes\! q_{\text{est}}$
  \State $\mathbf{P}\!\leftarrow\!(\mathbf{I}-\mathbf{K}\mathbf{H})\mathbf{P}$
\EndFor
\end{algorithmic}
\end{algorithm}

\end{minipage}
% }
\hfill
% \scalebox{0.9}{%
\begin{minipage}[t]{0.5\textwidth}
% \captionsetup{type=algorithm}
\begin{algorithm}[H]
\caption{Sliding-window FGO}
\label{alg:fgo_invpin}
\begin{algorithmic}[1]
\State \textbf{Node:} $\mathbf{T}_k=[x_k,y_k,z_k,\psi_k]^\top$, window $W$
\State Wait first absolute fix; add prior $\mathbf{r}_{\text{prior}}=\mathbf{T}_0-\hat{\mathbf{T}}_0$
\For {each confirmed step $k$}
  \State $\Delta\mathbf{u}_k=[\Delta x,\Delta y,\Delta z,\Delta\psi]^\top$
  \State Add node guess $\mathbf{T}_k \gets \mathbf{T}_{k-1}\oplus \Delta\mathbf{u}_k$
  \State Add Between: $\mathbf{r}_{\text{PDR}}=\mathrm{between}(\mathbf{T}_{k-1},\mathbf{T}_k)-\Delta\mathbf{u}_k$
  \If {absolute fix available}
    \State Add Position: $\mathbf{r}_{\text{abs}}=[x_k,y_k,z_k]^\top-\hat{\mathbf{p}}_k$
  \EndIf
  \If {nodes $>W$}
    \State Prune stale factors; weak-anchor oldest node
  \EndIf
  \State Optimize: $\min \sum\|\mathbf{r}_{\text{prior}}\|^2+\sum\|\mathbf{r}_{\text{PDR}}\|^2+\sum\|\mathbf{r}_{\text{abs}}\|^2$
\EndFor
\end{algorithmic}
\end{algorithm}

\end{minipage}
% }
\hfill
% \scalebox{0.9}{%
% ===================== BOTTOM ROW (1 centered block) =====================
% \makebox[\textwidth][c]{%
\begin{minipage}[t]{0.6\textwidth}
\begin{algorithm}[H]
% \captionsetup{type=algorithm}
\caption{Step-driven Particle Filter}
\label{alg:pf_invpin}
\begin{algorithmic}[1]
\State \textbf{Particle:} $\mathbf{x}^{(i)}=[x,y,z,\psi]^\top$, weights $w^{(i)}$, $N$
\State Init at first absolute ENU fix; $w^{(i)}\gets 1/N$
\For {each confirmed step $k$}
  \State $\Delta\mathbf{p}\!\gets\! \mathbf{p}^{\text{PDR}}-\mathbf{p}^{\text{PDR}}_{\text{prev}}$, 
         $\Delta\psi\!\gets\!\mathrm{wrap}(\psi^{\text{PDR}}-\psi^{\text{PDR}}_{\text{prev}})$
  \For {$i\gets 1\to N$}
    \State $\mathbf{p}^{(i)}\!\gets\!\mathbf{p}^{(i)}+\Delta\mathbf{p}+\boldsymbol{\epsilon}_p$
    \State $\psi^{(i)}\!\gets\!\mathrm{wrap}(\psi^{(i)}+\Delta\psi+\epsilon_\psi)$
  \EndFor
  \If {absolute fix $\hat{\mathbf{p}}$ available}
    \State $d_M^{2(i)}\!\gets\!(\mathbf{p}^{(i)}-\hat{\mathbf{p}})^\top \mathbf{R}^{-1}(\mathbf{p}^{(i)}-\hat{\mathbf{p}})$
    \State $w^{(i)}\!\gets\! w^{(i)}\exp(-\tfrac{1}{2}d_M^{2(i)})$; normalize
    \State $\mathrm{ESS}\!\gets\! 1/\sum_i (w^{(i)})^2$; resample if $\mathrm{ESS}<\tau N$
  \EndIf
  \State $\hat{\mathbf{x}}\!\gets\!\sum_i w^{(i)}\mathbf{x}^{(i)}$
\EndFor
\end{algorithmic}
\end{algorithm}

\end{minipage}
% }

\vspace{-0.4em}
\end{figure*}

\begin{figure}[t]
\centering

% Left column: two stacked subfigures
\begin{minipage}[b]{0.3\textwidth}
    \centering

    \subcaptionbox{Building polygons\label{fig:building_polygons}}{%
        \includegraphics[width=0.95\linewidth]{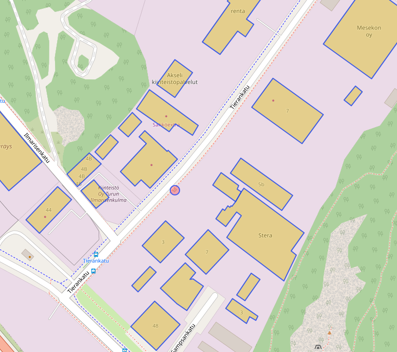}
    }

    \vspace{0.8em}

    \subcaptionbox{Map constraint\label{fig:map_constraint_demo}}{%
        \includegraphics[width=0.95\linewidth,trim=10 10 10 10, clip]{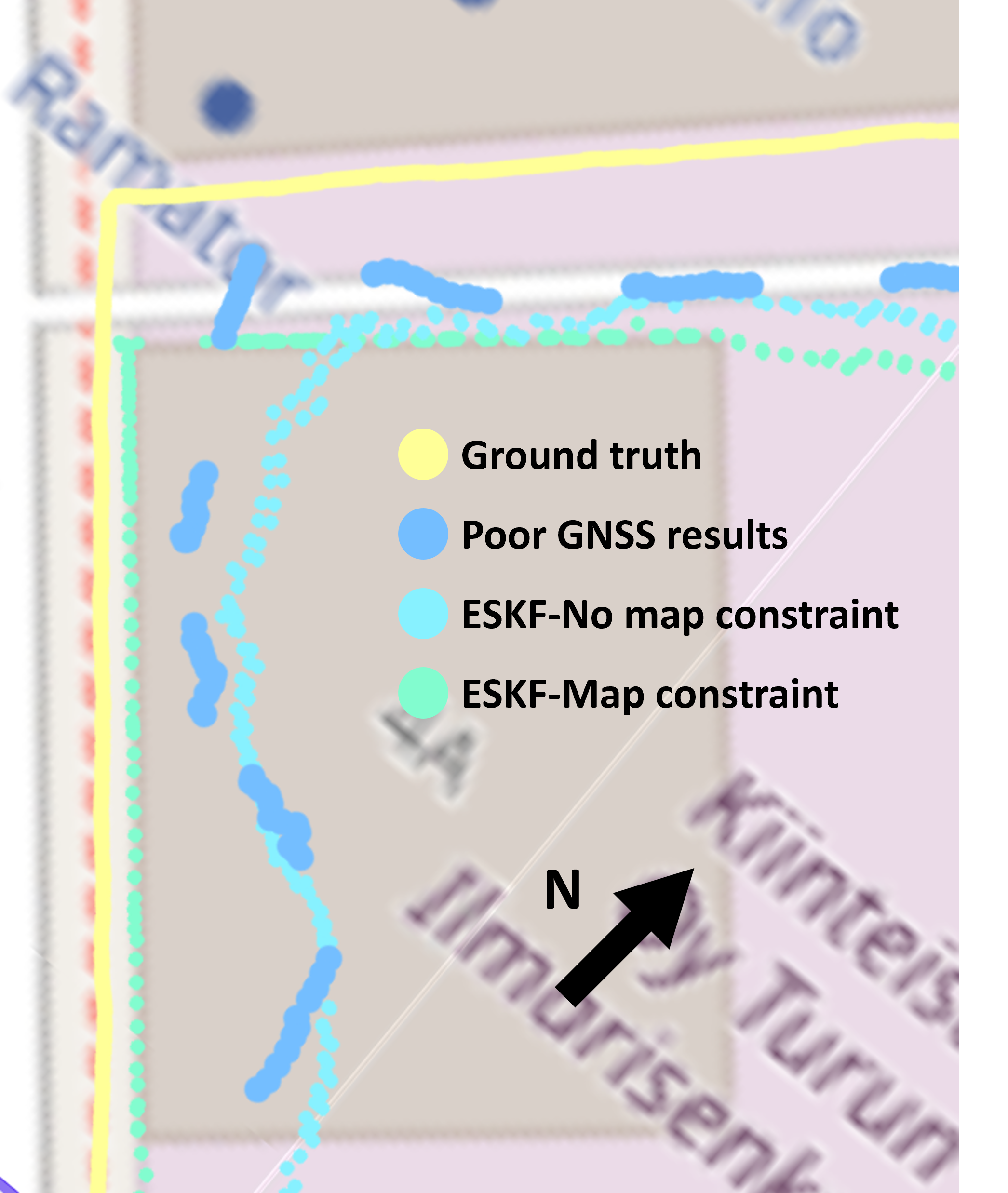}
    }
\end{minipage}
\hfill
% Right column: one subfigure
\begin{minipage}[b]{0.68\textwidth}
    \centering
    \subcaptionbox{Chest-Mounted Sensor Platform\label{fig:sensor_platform}}{%
        \includegraphics[width=0.95\linewidth, trim={200 50 250 10}, clip]{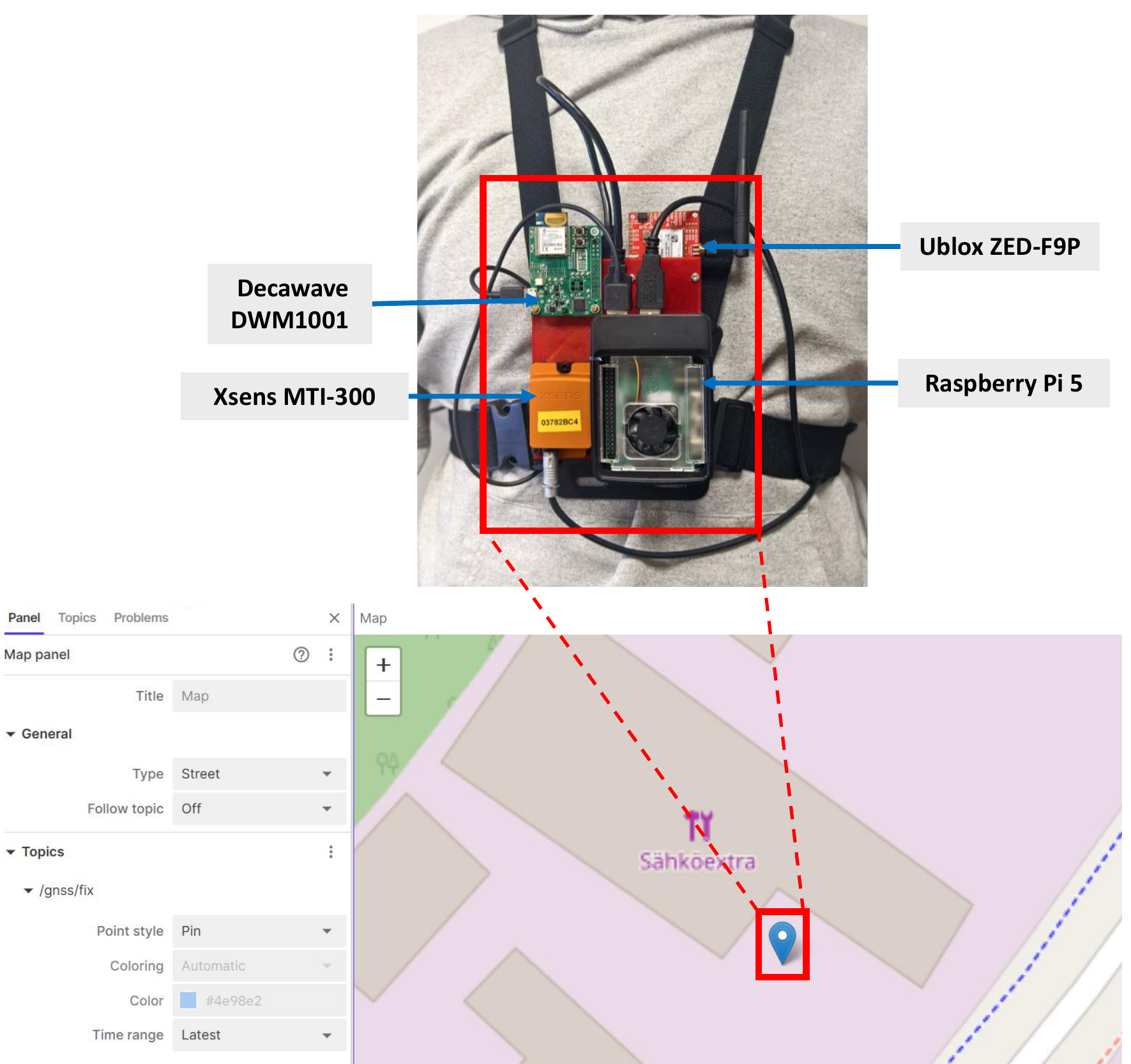}
    }
\end{minipage}

% Optional overall caption for the whole group
%     \caption{Chest-Mounted Sensor Platform}
    \label{fig:platform_overview}
\caption{Overview of map constraints and the chest-mounted sensor platform.}
% \vspace{-10pt}
\end{figure}

\subsection{Experimental Setting and Visualization}
\label{subsec:exp_setting}
The wearable platform integrates an MTi-300 IMU, a GNSS receiver, and a DWM1001 UWB tag on a rigid chest-mounted plate driven by a Raspberry Pi~5, as shown in Fig.~\ref{fig:sensor_platform}. 
All data are timestamped and logged via ROS\,2, and the three fusion back-ends publish their outputs using consistent message interfaces for fair comparison.

Experiments are conducted on a campus site that includes outdoor walkways and an adjacent instrumented laboratory building. We evaluate three scenarios: (i) outdoor walks with GNSS+PDR, (ii) indoor walks with UWB+PDR, and (iii) repeated outdoor--indoor transitions with both absolute sources available intermittently. Outdoor reference is provided by an RTK-capable solution, while the indoor reference is provided by a motion-capture system aligned to the same local ENU frame.

For efficient online inspection and offline figure generation, a unified Foxglove dashboard visualizes raw GNSS, UWB positions, three fused trajectories (ESKF/FGO/PF), PDR step events, and essential diagnostic signals. This setup allows us to trace estimator behavior around challenging segments such as GNSS-degraded building proximities and UWB anchor-visibility changes.

\section{Experimental Results}\label{sec:experiment}

% \subsection{Pedestrian Dead Reckoning with Chest-mounted IMU}
We evaluate three scenarios: indoor with IMU and UWB using motion capture (MoCap) as ground truth, outdoor with GNSS (u-blox ZED-F9P) and IMU using RTK as ground truth, and a seamless outdoor–indoor experiment with all three sensors where the MoCap system publishes latitude–longitude and the ground truth is the fusion of RTK when outside and MoCap when inside. For all scenarios, accuracy is assessed against ground truth using the horizontal error and reported as cumulative distribution functions (CDFs), shown in Fig.~\ref{fig:all-cdf},  together with summary statistics, shown in Table~\ref{tab:indoor_outdoor_metrics}. 

The pedestrian walked a square loop three times inside the allowed building. 
As shown in Fig.~\ref{fig:cdf_indoor} and Table~\ref{tab:indoor_outdoor_metrics}, ESKF achieves the best indoor accuracy (median 0.415~m, RMSE 0.499~m), closely followed by PF (median 0.447~m, RMSE 0.552~m), while FGO shows larger errors. 

The outdoor trial covers sidewalks and building façades with GNSS/PDR fusion. 
ESKF again provides the most accurate distribution (median 1.491~m, RMSE 2.186~m), followed by FGO and PF (Table~\ref{tab:indoor_outdoor_metrics}). 
The building-polygon constraint mainly improves robustness against multipath outliers, which is also reflected by cleaner tails in the CDF (Fig.~\ref{fig:cdf_outdoor}).

\begin{figure}[t]
    \centering
    \begin{subfigure}{0.32\textwidth}
        \includegraphics[width=0.8\textwidth]{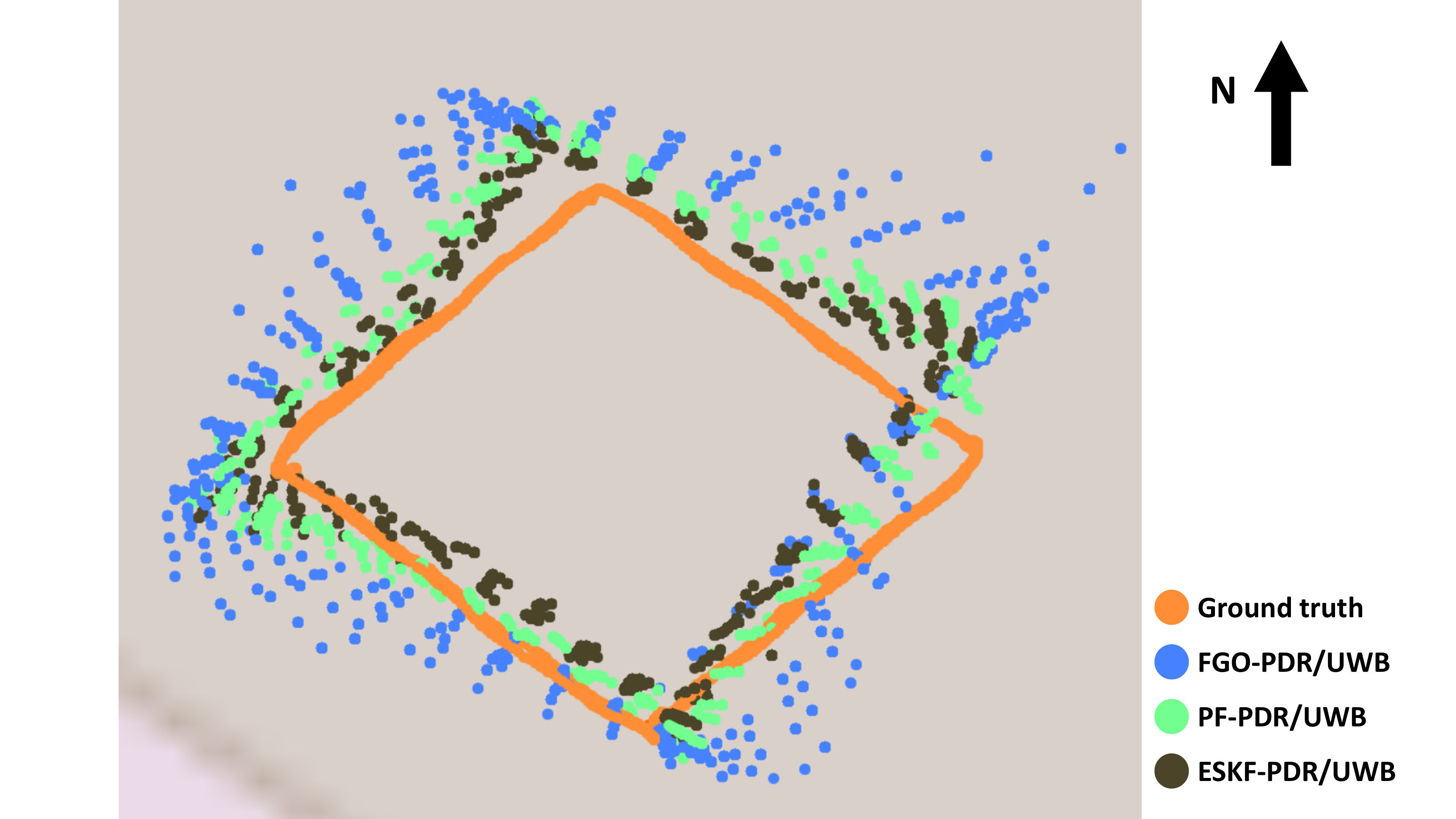}
        % \includestandalone[width=0.95\textwidth]{fig/outdoor_err2d_cdf}
        \caption{Indoor trajectory}
        \label{fig:traj_indoor}
    \end{subfigure}
    \begin{subfigure}{0.32\textwidth}
        \includegraphics[width=0.80\textwidth,trim=10 20 10 10, clip]{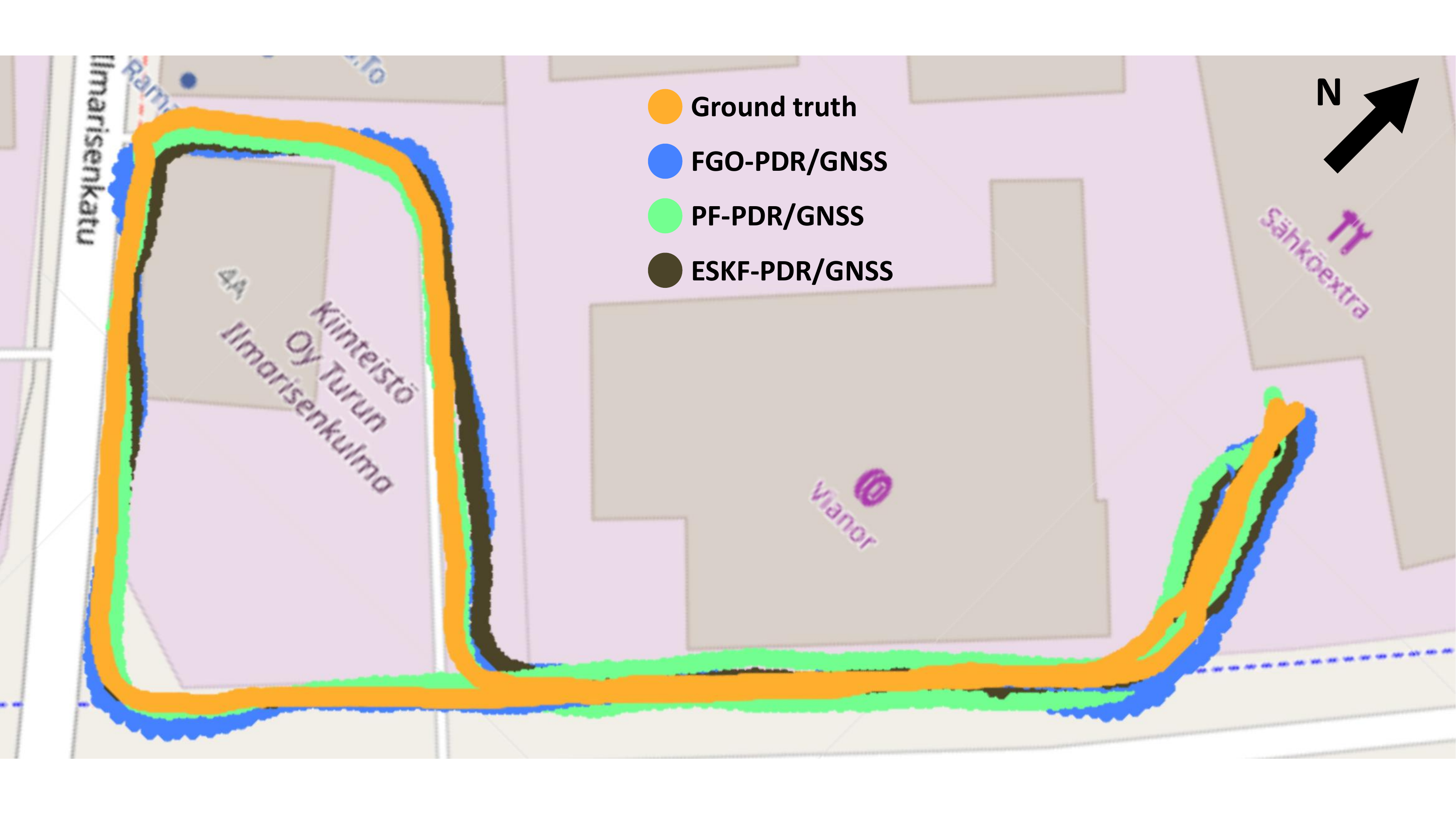}
        % \includestandalone[width=0.95\textwidth]{fig/outdoor_err2d_cdf}
        \caption{Outdoor trajectory}
        \label{fig:traj_outdoor}
    \end{subfigure}
    \begin{subfigure}{0.32\textwidth}
        \includegraphics[width=0.80\textwidth]{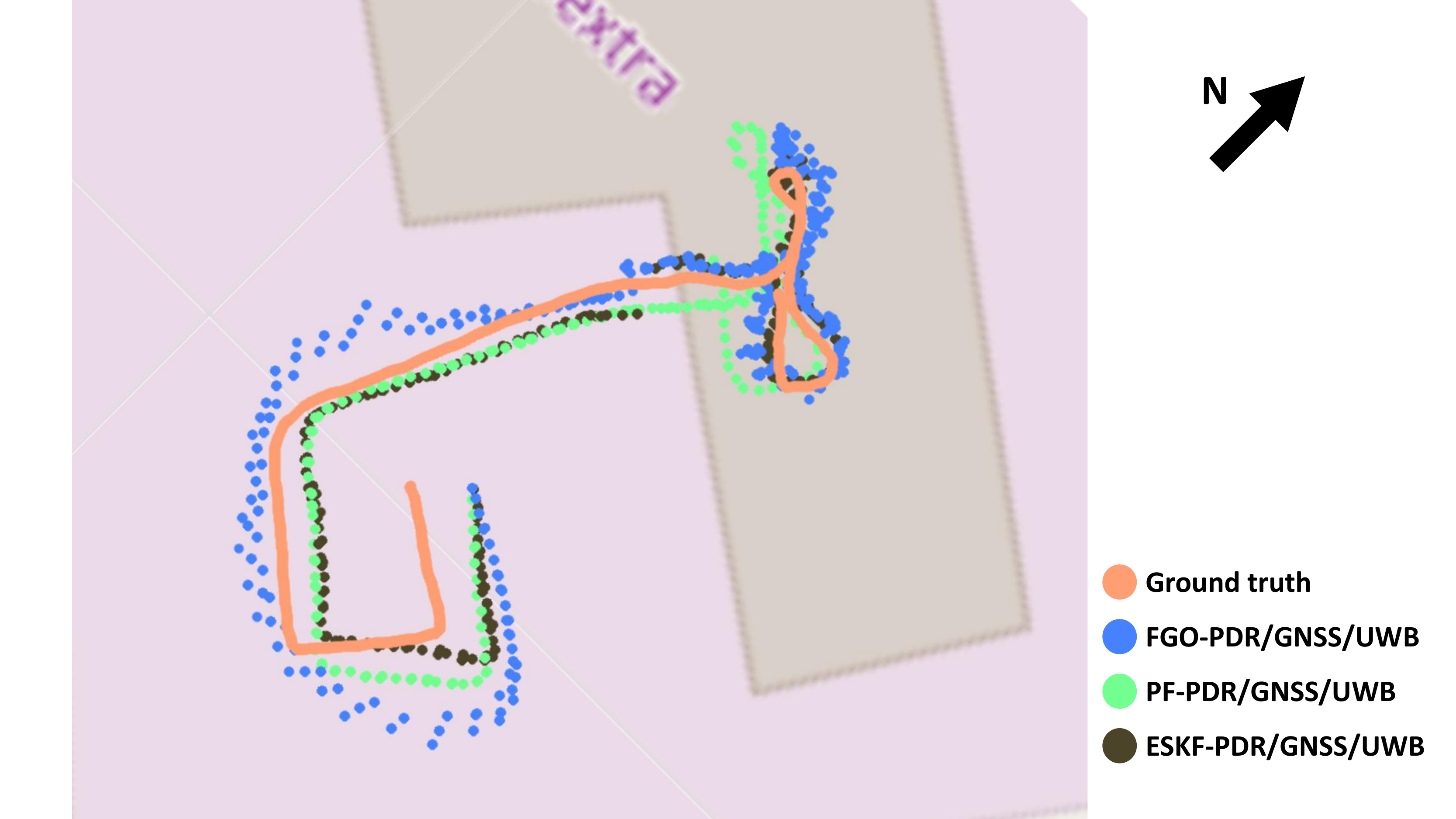}
        % \includestandalone[width=0.95\textwidth]{fig/outdoor_err2d_cdf}
        \caption{Outdoor to indoor trajectory}
        \label{fig:traj_outdoor_indoor}
    \end{subfigure}
    \caption{Trajectories of the compared methods under different environmental settings.}
    \label{fig:traj}
    \vspace{-10pt}
\end{figure}

In the seamless trial, the estimator naturally transitions from GNSS to UWB updates at the doorway without visible discontinuities in the estimated trajectory (Fig.~\ref{fig:traj_outdoor_indoor}). 
Across the full sequence, ESKF yields the best overall accuracy (median 0.997~m, RMSE 1.248~m), PF is second, and FGO is less stable in this mixed regime (Table~\ref{tab:indoor_outdoor_metrics}). 
These results indicate that a tightly managed prediction--update cycle with consistent covariance handling (ESKF) is particularly effective for maintaining continuity across heterogeneous sensing conditions.

\begin{table}[H]
\centering
\caption{Indoor/Outdoor experiment}
\label{tab:indoor_outdoor_metrics}
\resizebox{0.9\textwidth}{!}{
\begin{tabular*}{\hsize}{@{\extracolsep{\fill}}lccc@{}}
\toprule
\multirow{2}{*}{}        & \multicolumn{1}{c}{\textbf{Indoor}} & \multicolumn{1}{c}{\textbf{Outdoor}} & \multicolumn{1}{c}{\textbf{Outdoor-Indoor}} \\
                         & \multicolumn{3}{c}{\textit{\textbf{(Mean, Median, RMSE. STD, Max), Unit: m}}}                                            \\
\colrule
\textbf{FGO}             & (0.836, 0.715, 0.992, 0.535, 3.303)     & (1.995, 1.946, 2.389, 1.315, 7.619)      & (2.070, 1.962, 2.302, 1.008, 5.351)             \\
\textbf{Particle filter} & (0.491, 0.447, 0.552, 0.253, 1.374)     & (2.284, 1.681, 2.816, 1.648, 7.357)      & (1.292, 1.017, 1.653, 1.031, 5.108)             \\
\textbf{ESKF}            & (0.441, 0.415, 0.499, 0.235, 1.309)     & (1.726, 1.491, 2.186, 1.341, 8.181)      & (1.085, 0.997, 1.248, 0.617, 2.631) \\  
\botrule
\end{tabular*}
}
\vspace{-10pt}
\end{table}

\begin{figure}[t]
    \centering
    \begin{subfigure}{0.32\textwidth}
        \centering
        \includestandalone[width=1.0\textwidth]{fig/indoor_cdf}
        \caption{Indoor results}
        \label{fig:cdf_indoor}
    \end{subfigure}
    \begin{subfigure}{0.32\textwidth}
        \centering
        \includestandalone[width=\textwidth]{fig/outdoor_cdf_downsample}
        \caption{Outdoor results }
        \label{fig:cdf_outdoor}
    \end{subfigure}
    \begin{subfigure}{0.32\textwidth}
        \centering
        \includestandalone[width=\textwidth]{fig/outdoor_indoor_cdf_downsample}
        \caption{Outdoor indoor results}
        \label{fig:cdf_outdoor_indoor}
    \end{subfigure}
    \caption{The comparison of CDF under different experimental settings.}
    \label{fig:all-cdf}
    \vspace{-0.1cm}
\end{figure}

\section{Conclusion}\label{sec:conclusion}
This paper presented a seamless outdoor--indoor pedestrian positioning system that fuses GNSS/UWB/IMU under a unified chest-mounted PDR prediction framework, and systematically compared three estimators: ESKF, FGO, and PF. We also incorporated lightweight map-derived building polygon constraints to improve feasibility and robustness in proximity to buildings and during indoor--outdoor transitions.

Across indoor, outdoor, and outdoor--indoor experiments, ESKF delivered the most consistent and accurate performance overall. Indoors, where UWB provides absolute corrections to PDR, ESKF achieved the best error distribution, with PF as a close runner-up, while FGO exhibited larger residual errors in this setup. Outdoors, ESKF remained the most stable against RTK-referenced ground truth, and the map constraint effectively suppressed multipath-driven outliers near façades. In the seamless trial, the estimator naturally shifted from GNSS to UWB updates at the doorway without noticeable trajectory discontinuities, and ESKF again provided the best overall accuracy. These results indicate that covariance-consistent prediction--update management in ESKF is particularly advantageous under heterogeneous sensing availability and quality.

Overall, the findings suggest that an ESKF-based fusion pipeline with PDR-driven propagation, GNSS/UWB opportunistic updates, and modest map feasibility checks offers a strong balance of accuracy, robustness, and implementation simplicity for practical pedestrian navigation across indoor--outdoor boundaries. PF remains a competitive alternative when measurement uncertainty is heavy-tailed or difficult to model, while FGO may require richer factor designs and more careful sensor-characteristic tuning to match filter-based performance in this specific configuration.

% \paragraph{Limitations and Future Work}
% Future work will focus on (i) enhancing the FGO formulation with more informative motion and measurement factors (e.g., adaptive UWB NLOS handling and transition-aware priors), (ii) improving PF efficiency via better proposal distributions and adaptive resampling, and (iii) extending the map constraint from simple feasibility gating to probabilistic, semantics-aware constraints that can exploit entrances, corridors, and multi-floor structures. We also plan to evaluate the system under denser crowds, longer trajectories, and more diverse building layouts to further validate generalizability.

\section*{Acknowledgements}
% $ $ \\[-30pt]
This research work is supported by the Research Council of Finland's Digital Waters (DIWA) flagship (Grant No. 359247), DIWA Doctoral Training Pilot project funded by the Ministry
of Education and Culture (Finland), as well as the R3Swarms project funded by the Secure Systems Research Center (SSRC), Technology Innovation Institute (TII).

\bibliographystyle{elsarticle-num}
\bibliography{bibliography.bib}

\end{document}